\documentclass[11pt]{article}

\usepackage[utf8]{inputenc}
\usepackage[T1]{fontenc}
\usepackage[english]{babel}
\usepackage{lmodern}
\usepackage{microtype}
\usepackage[margin=1in]{geometry}
\usepackage{setspace}
\usepackage{amsmath,amsfonts,amssymb,bm}
\usepackage{graphicx}
\usepackage{float}
\usepackage{subcaption}
\usepackage{pgfplots}
\usepackage{booktabs}
\usepackage{threeparttable}
\usepackage[table]{xcolor}
\usepackage{soul}
\usepackage{caption}
\usepackage{titlesec}
\usepackage{authblk}
\usepackage[numbers,sort&compress]{natbib}
\usepackage[hidelinks]{hyperref}

\pgfplotsset{compat=1.18}
\titleformat{\section}{\large\bfseries}{\thesection}{0.75em}{}
\titleformat{\subsection}{\normalsize\bfseries}{\thesubsection}{0.75em}{}
\titleformat{\subsubsection}{\normalsize\itshape}{\thesubsubsection}{0.75em}{}
\titlespacing*{\section}{0pt}{1.5em}{0.6em}
\titlespacing*{\subsection}{0pt}{1.2em}{0.4em}
\titlespacing*{\subsubsection}{0pt}{1em}{0.3em}

\title{\bfseries Deep Learning-Based Pavement Performance Modeling Using Multiple Distress Indicators and Road Work History}
\author[1]{Lu Gao\thanks{Corresponding author. Email: \href{mailto:lgao5@central.uh.edu}{lgao5@central.uh.edu}}}
\author[2]{Zhe Han}
\author[3]{Yunshen Chen}
\affil[1]{Department of Construction Management, University of Houston}
\affil[2]{Center for Transportation Research, The University of Texas at Austin}
\affil[3]{Independent Scholar}

\date{}

\begin{document}

\maketitle
\begin{abstract}
Pavement deterioration is a complicated process influenced by various factors such as design, environment, material, and other unobserved variables. Reliable and accurate predictions of pavement condition can save significant amounts of resource for pavement management agencies through better planned maintenance and rehabilitation activities. In this paper, the authors employed deep learning networks of convolutional neural network (CNN), long short-term memory (LSTM), and a CNN-LSTM combination to capture the pavement deterioration rate. While traditional models are limited by their ability to handle raw data, deep learning models do not need the conventional steps of feature extraction and feature selection. These steps are embedded in the deep learning framework through a self-learning process. In this paper, pavement condition data and maintenance and rehabilitation history collected by the Texas Department of Transportation in the recent 18 years were used. Twenty-one flexible pavement condition indicators, including cracking, rutting, raveling, and roughness, collected from more than 100,000 pavement sections were used to develop the proposed models. Promising preliminary results were obtained.
\end{abstract}


\section{Introduction}

Pavement deterioration is a complicated process influenced by various factors such as design, environment, material, and other unobserved variables \citep{gao2019impacts}. The natural deterioration of the existing highway systems has resulted in the expenditure of a large portion of highway funds on pavement maintenance and rehabilitation \citep{han2019incorporating,politis2020stochastic}. In addition, over the past decades, the traffic volumes on primary highway systems have experienced tremendous increases, which accelerates deterioration and leads in many instances to premature failures of highway pavements \citep{chen2003forensic, ziari2007interface}. An accurate and complete understanding of pavement performance and deterioration process can not only predict pavement performance over time, but also is essential for a successful pavement management system (PMS), which further assists transportation agencies to optimize their highway maintenance and rehabilitation programs by allocating budgets more reasonably \citep{xu2021development}. 

Various deterioration models (or performance models) are developed based on historical condition inspection data \citep{gao2022missing,yu2023pavement}. A popular example of models used by highway agencies is based on the Markov Chain theory, which are widely adopted in practice because such models require less frequent data collection. The core of the Markov Chain models is the development of the transition probabilities, representing the chances of a pavement deteriorating from one condition state to another. \citet{golabi1982statewide} proved the effectiveness of using the Markov Chain method by developing pavement Markov Chain performance models in Arizona. A number of methods, including the expected-value method by \citet{butt1994application} and \citet{jiang1988bridge} and the proportion method by \citet{wang1994probabilistic}, have been employed to develop transition probabilities. \citet{li1996reliability} presented a reliability-based processing of Markov Chains for modeling pavement network deterioration by applying Monte Carlo Simulation technique. \citet{black2005semi} proposed a semi-Markov approach for modelling asset deterioration by relaxing the assumption that constant transition probabilities are irrespective of how long an item has been in a state. \citet{ortiz2006derivation} used historical data, regression curve, and yearly distribution of pavement condition to derive transition probability matrices for pavement deterioration modeling. \citet{gao2007using} proposed a simulation approach of utilizing design equations to developing transition probabilities. \citet{kobayashi2012statistical} adopted exponential hazard models to estimate the Markov transition probability model to forecast the deterioration process of road sections. \citet{kobayashi2012statistical} presented a hidden Markov model to tackle selection biases in monitoring data to forecast road deterioration. \citet{thomas2013comparison} applied Weibull distribution present a semi-Markov model to investigate pavement deterioration patterns. Using pavement condition data, \citet{perez2018transition} demonstrated the feasibility to develop transition probability matrices for an entire flexible pavement network and carry out maintenance and rehabilitation activities simultaneously. 

Besides Markov Chain, other traditional statistical models with different functional forms were developed to model pavement deterioration \citep{jahanbakhsh2016estimating}. \citet{madanat1995estimation} applied two econometric methods to estimate joint discrete-continuous models of pavement distress initiation and progression while correcting for selectivity bias. \citet{madanat1998development} jointed a discrete model of distress initiation and a continuous model of pavement progression to develop a pavement distress progression model using panel data sets of in-service pavements. \citep{prozzi2000using} used duration models to analyze experimental pavement failure data to estimate pavement conditions. \citet{prozzi2004development} combined experimental and field data to develop a pavement performance model. \citet{li2005probabilistic} utilized ordered probit model and sequential logit model to capture the dynamic and stochastic nature of pavement deterioration processes. \citet{wang2005survival} performed survival analysis of fatigue cracking for flexible pavements to investigate the relationship between pavement fatigue failure time and various influencing factors. \citet{zhang2006applying} and \citet{han2019incorporating} combined reliability theory and method of moments to model pavement deterioration while incorporating various uncertainties associated with a pavement. \citet{wang2008dynamic} employed a dynamic panel data prediction method to estimate the performance of asphalt concrete overlay. \citet{chu2008empirical} presented state-space specifications of time series models to formulate dynamic performance models for pavements and estimated them using panel data sets. 
\citet{hong2010roughness} developed a Bayesian nonlinear pavement deterioration model based on data from in-service pavement sections. \citet{gao2011performance} proposed a Bayesian-based robust performance model that can calculate the probability that a data point is affected by maintenance intervention. \citet{hong2015pavement} developed a nonlinear pavement deterioration model, which integrated construction, design, structure, material, time and traffic, environment, and maintenance. Based on discrete choice model theory, \citet{zhang2016nested} developed a probabilistic model to capture the stochastic nature of pavement deterioration process and predict the probability of a pavement staying at defined condition states. \citet{pantuso2019development} applied negative binomial regression and linear empirical Bayesian technique to develop network-level pavement deterioration curves. 

Machine learning models have also been used in pavement and other infrastructure performance and deterioration modeling. \citet{attoh1994predicting} predicted roughness progression in flexible pavements using supervised learning artificial neural network (ANN). \citet{owusu1998application} applied ANN to model thick asphalt pavement performance. \citet{attoh1999analysis} developed a flexible pavement deterioration model using back-propagation type of ANN and investigated investigate the effect of learning rate and momentum term based on real pavement data. \citet{choi2004pavement} adopted a back-propagation neural network algorithm to model pavement roughness. \citet{morcous2005prediction} integrated ANN, case-based reasoning, mechanistic model, and Monte Carlo simulation to propose a probabilistic mechanistic model for modeling the performance of reinforcing steel in concrete bridge decks. \citet{yang2008automated} adopted three neural network approaches, back-propagation neural network (BPN), radial basis network (RBN), and support vector machine (SVM), to classify sewer pipe defect patterns. \citet{thube2012artificial} developed pavement deterioration models based on ANN to forecast cracking, raveling, rutting, and roughness for low volume roads in India. \citet{tabatabaee2013two} used a support vector classifier (SVC) and a recurrent neural network (RNN) to classify and accurately predict the performance of a pavement infrastructure system. \citet{gajewski2014sensitivity} conducted sensitivity analysis of crack propagation in pavement bituminous layered structures using a hybrid system integrating ANN and finite element method. \citet{kirbacs2016performance} developed deterministic regression, multivariate adaptive regression splines, and ANN deterioration models for HMA paved road sections in urban roads, and the ANN method was found to be the most appropriate model for predicting deterioration. \citet{barua2020gradient} presented two gradient boosting approaches to estimate pavement deteriorations of airport runways and taxiways. They found that the developed models are shown to outperform other methods (including linear regression, nonlinear regression, artificial neural networks, and random forest) in terms of model goodness-of-fit for both runway and taxiway pavements. 

More recently, with the development of big data and artificial intelligence, deep learning models have been applied to the pavement performance modeling area. Compared with traditional models, deep learning models don't need the conventional steps of feature extraction and can be applied to raw data directly. Recent research shows that deep learning models (e.g., LSTM) have been adopted in modeling pavement performance and deterioration processes due to better performance and higher accuracy. For example, \citet{lee2019development} developed a pavement deterioration prediction model based on deep neural network and RNN with Long short-term memory circuits (LSTM). They found that the performance and accuracy of the LSTM model was superior. \citet{choi2020development} predicted the deterioration of road pavement by using monitoring data and a LSTM framework. The constructed algorithm predicts the pavement condition index for each section of the road network for one year by learning from the time series data for the preceding 10 years. In this paper, we extended the previous applied LSTM framework by adding the Convolutional Neural Network model to evaluate their performance on pavement condition data collected from Texas road network. 

The remaining sections are organized as follows: Section 2 introduces the concepts of CNN, LSTM, and CNN-LSTM techniques in details; Section 3 includes the empirical case study and the results are analyzed, compared, and interpreted; Section 4 summarizes the results of this paper, and conclusions are drawn by verifying the applicability of the proposed CNN-LSTM algorithm in predicting pavement performance.

\section{Methodology}
In this paper, We have adopted deep learning techniques for the detection of pavement maintenance treatments. Traditional machine learning models are limited by their ability to handle raw data. The advantage of deep learning model is that the conventional steps of feature extraction and feature selection are no longer needed. These steps are embedded in the deep learning framework through self-learning process \citep{lecun2015deep}. In the following sections, we will discuss three deep learning models used in the paper: convolutional neural network model (CNN), long short-term memory (LSTM) model, and a hybrid CNN-LSTM model.

\subsection{Notations}
We use a symbol shown in boldface to represent a vector, e.g., $ \bm{x} \in \mathbb{R}^D$ is a column vector with $D$ elements. We use a boldface capital letter to denote a matrix, e.g., $\bm{X}\in \mathbb{R}^{H\times W}$ is a matrix with $H$ rows and $W$ columns. Scalar variables are denoted by ordinary letters. The variables mentioned in this paper are summarized in Table \ref{tab:notations}.

\begin{table}[H]
\caption{\label{tab:notations}List of Symbols}
\begin{tabular}{|c|p{10cm}|}
    \hline
    \textbf{Variables} & \textbf{Definition}\\
    \hline
    $\bm{c}_{t}^{i}$ & Memory or cell unit at time $ t $ of the $ i $th layer\\    
    $\bm{f}_{t}^{i}$ & Forget gate at time $ t $ of the $ i $th layer\\
    $\bm{h}_{t}^{i}$ & Hidden state at time $ t $ of the $ i $th layer\\    
    $\bm{g}_{t}^{i}$ & Input modulation gate at time $ t $ of the $ i $th layer\\
    $\bm{k}_{t}^{i}$ & Input gate at time $ t $ of the $ i $th layer\\
    $K$ & Total number of layers\\    
    $\bm{o}_{t}^{i}$ & Output gate at time $ t $ of the $ i $th layer\\
    $\sigma(\cdot)$ & Sigmoid function\\    
    $T$ & Total number of time steps\\       
    $\bm{w}^i$ & Vector of parameters (to be estimated) of the $ i $th layer (including $\bm{w}_{xf}^i$, $\bm{w}_{hf}^i$, $\bm{b}_{f}^i$, $\bm{w}_{xk}^i$, $\bm{w}_{hk}^i$, $\bm{b}_{k}^i$, $\bm{w}_{xg}^i$, $\bm{w}_{hg}^i$, $\bm{b}_{g}^i$, $\bm{w}_{xo}^i$, $\bm{w}_{ho}^i$, and $\bm{b}_{o}^i$)\\
    $\bm{x}^i_t$ & Input data of the $t^{th}$ timestep in the $ i $th layer\\    
    $\bm{X}^i$ & Input data of the $ i $th layer\\        
    $\otimes$ & Pointwise multiplication \\
    $\oplus$ & Pointwise addition \\    
    \hline
\end{tabular}
\end{table}

\subsection{Convolutional neural network}
The convolutional neural network was first proposed by \cite{lecun1990handwritten}. It is a special type of feedforward neural network (FFNN). Inspired by the “neocognitron”  model \citep{fukushima1979neural}, convolutional neural network (CNN) was initially used in document recognition with the implementation of the “LeNet” model \citep{lecun1998gradient}. In 2012, a CNN  model named “AlexNet” was invented \citep{krizhevsky2012imagenet}, which opened the gate of using CNN for imaging process applications. A few years later, Simonyan and Zisserman introduced the “VGG16” model for imaging recognition with very deep convolutional networks. Compared with traditional fully connected neural networks, CNN utilizes shared parameters among features and sparsity of connections, which allow us to encode arbitrary large items. Such properties also enable CNN to capture indicative local predictors in a large structure and combine the local information for a fixed size vector representation of the structure. Recently, CNN has been used in the regions outside imaging processing  such as time series classification tasks \citep{cui2016multi, wang2017time} and showed promising performance, which inspires us to apply it on time series data in transportation discipline. 

\begin{figure}[H]
	\includegraphics[width=1\textwidth]{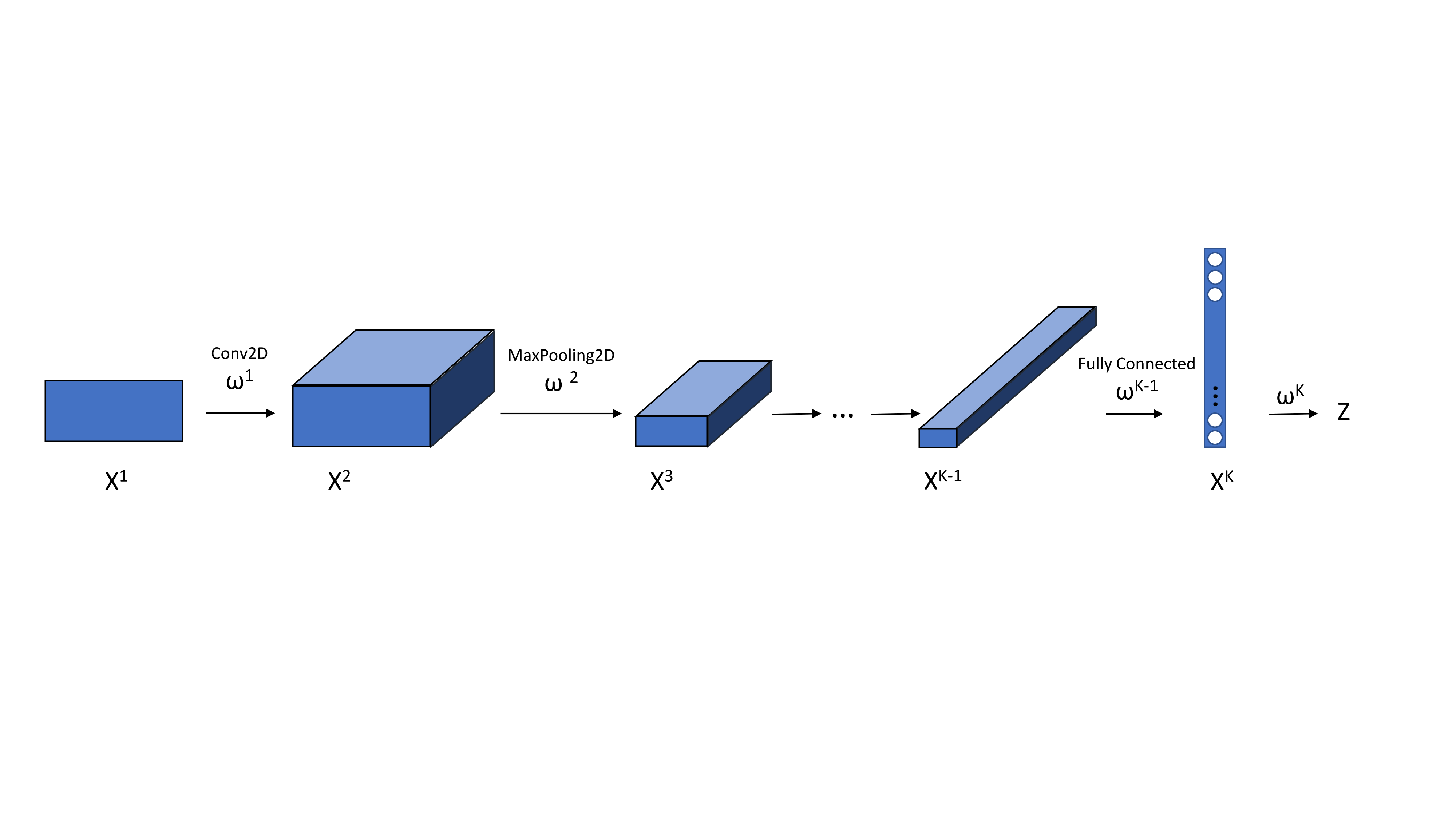}
	\caption{\label{fig:cnn_structure}CNN Structure}
\end{figure}

In this paper, each input data is a two-dimensional matrix $\bm{X}=(\bm{x}_1,\bm{x}_2,...,\bm{x}_{T})$ where 
$\bm{x}_t$ denotes pavement conditions (e.g., rutting, cracking, roughness) at time step $t$. An abstract CNN structure is given in Figure \ref{fig:cnn_structure}.


The above equation shows a CNN layer by layer in a forward format. The layers are labeled as boxes. The input $ \bm{X}^1 $ goes through the first layer denoted as $ \bm{w}^1 $, which represents a vector of parameters involved in the first layer's calculation. The output of the first layer is $ \bm{X^2} $, which is also the input to the second layer. The final output of the network is $ \bm{X}^K $. The last layer, $ \bm{w}^K $, is the loss layer, where a loss function is defined. 

\subsection{Long Short-Term Memory (LSTM)}

\begin{figure}[H]
\includegraphics[width=0.8\textwidth]{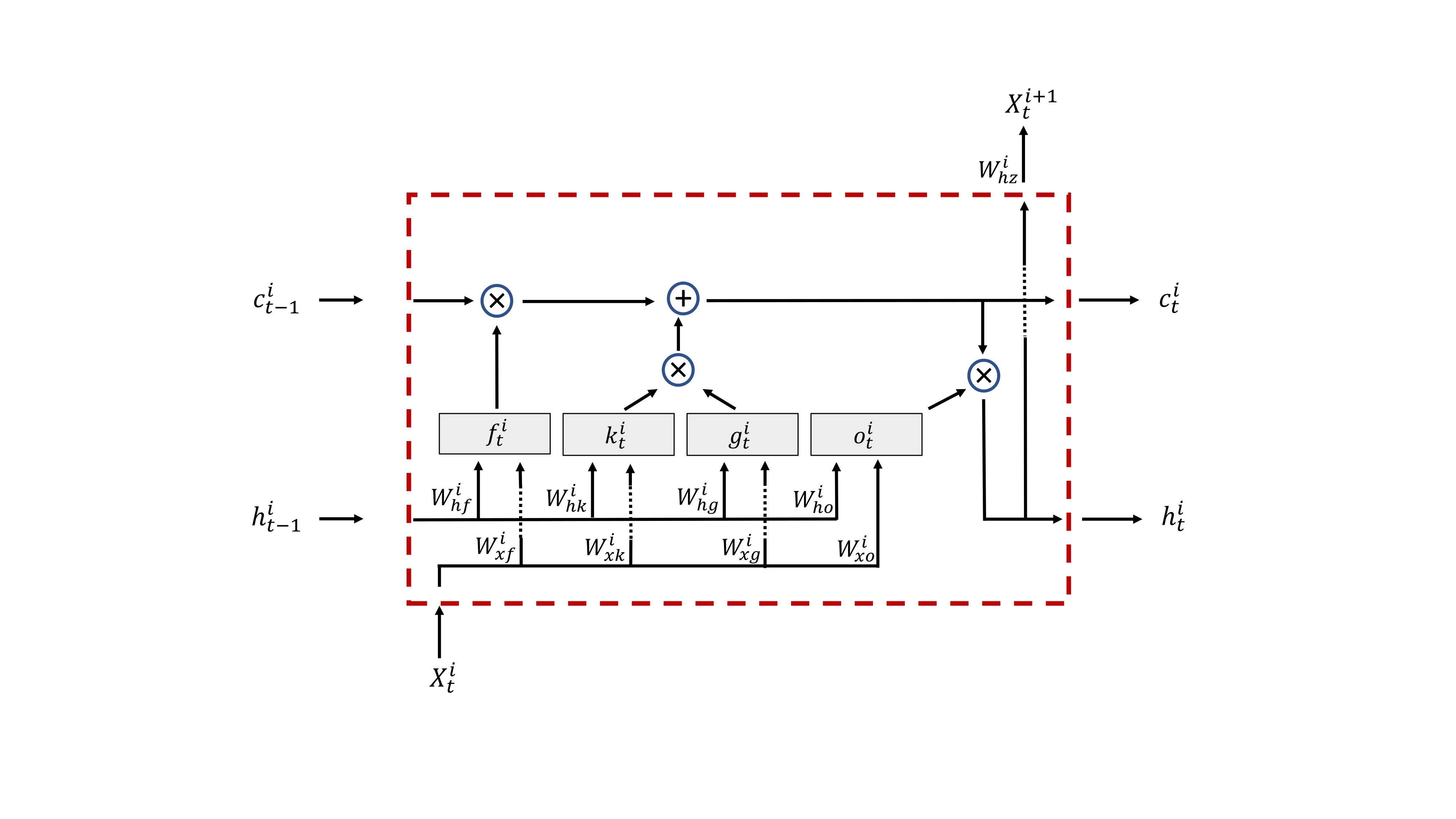}
\caption{\label{fig:lstm_structure}LSTM Block Structure in Unfolded Form}
\end{figure}

CNN is useful for classification tasks expected with strong local signals. However, the trade-off is the sacrifice of the structural information \citep{goldberg2016primer}. Recurrent neural network(RNN), in contrast, is good at preserving structural information. Particularly, a type of RNN model, named the Long short term memory (LSTM)  \citep{hochreiter1997long}, which was originally designed to solve the vanishing gradients problem encountered in traditional deep neural networks. The LSTM model introduces a “memory” cell- a gradient vector, and practices a logical gate system to decide whether the “memory” is going to be “forgot”, “updated” and “incorporated”  in an output prediction.  Recently, LSTM is attractive in time-series modeling, given its convenience in automatic feature engineering \citep{assaad2008new} and capability of capturing complicated feature interactions \citep{ogunmolu2016nonlinear}. The LSTM block is a complex process with many units (Figure \ref{fig:lstm_structure}). Compared with RNN, an extra memory unit, $\bm{c}_t^i$, is defined to store the information across many time steps with the control from several adaptive gating units. 

The forget gate, $f_t^i$, decides what cell unit information, $\bm{c}^i_t$, should be thrown away or kept through the $\otimes$ operation. Information from the previous hidden state, $\bm{h}^i_{t-1}$, and information from the current input, $\bm{x}_{t}^i$, is passed through the sigmoid function (Equation \ref{eq:forget}) together with weight matrices, $\bm{w}_{xf}^i$ and $\bm{w}_{hf}^i$, and bias vectors $\bm{b}_f^i$. If the result is closer to 0 it means to forget, otherwise it means to keep.

\begin{equation}\label{eq:forget}
\bm{f}_t^{i} = \sigma(\bm{w}^{i}_{xf}\bm{x}^{i}_{t}+\bm{w}^{i}_{hf}\bm{h}^{i}_{t-1}+\bm{b}^{i}_{f})
\end{equation}

The input gate, $\bm{k}_t^i$, and the input modulation gate, $\bm{g}^i_t$, are designed to update the cell unit $\bm{c}^i_{t}$. First, the previous hidden state and current input are used to calculate the input gate through a sigmoid function (Equation \ref{eq:input}). The result decides which values will be updated. 

\begin{equation}\label{eq:input}
\bm{k}_t^{i} = \sigma(\bm{w}^{i}_{xk}\bm{x}^{i}_{t}+\bm{w}^{i}_{hk}\bm{h}^{i}_{t-1}+\bm{b}^{i}_{k})
\end{equation}

Then the hidden state and current input will be put into the tanh function (Equation \ref{eq:input_mod}) to produce values between -1 and 1. The purpose of the tanh function is to help regulate the values.

\begin{equation}\label{eq:input_mod}
\bm{g}_t^{i} = tanh(\bm{w}^{i}_{xg}\bm{x}^{i}_{t}+\bm{w}^{i}_{hg}\bm{h}^{i}_{t-1}+\bm{b}^{i}_{g})
\end{equation}

The cell unit is updated through Equation \ref{eq:cell} by adding results from the input gate and the forgot gate together. This operation is to ensure that only important information is added to the cell unit. 

\begin{equation}\label{eq:cell}
\bm{c}_t^{i} = \bm{f}^{i}_{t}\bm{c}^{i}_{t-1}+\bm{k}^{i}_{t}\bm{g}^{i}_{t}
\end{equation}

The output gate $\bm{o}_t^i$ decides what the next hidden state should be. It is calculated through a sigmoid function (Equation \ref{eq:output}) and a tanh function (Equation \ref{eq:ht}). 

\begin{equation}\label{eq:output}
\bm{o}_t^{i} = \sigma(\bm{w}^{i}_{xo}\bm{x}^{i}_{t}+\bm{w}^{i}_{ho}\bm{h}^{i}_{t-1}+\bm{b}^{i}_{o})
\end{equation}

\begin{equation}\label{eq:ht}
\bm{h}_t^{i} = \bm{o}^{i}_{t}tanh(\bm{c}_t^i)
\end{equation}

\subsection{CNN-LSTM Model}

\begin{figure}[H]
	\includegraphics[width=1\textwidth]{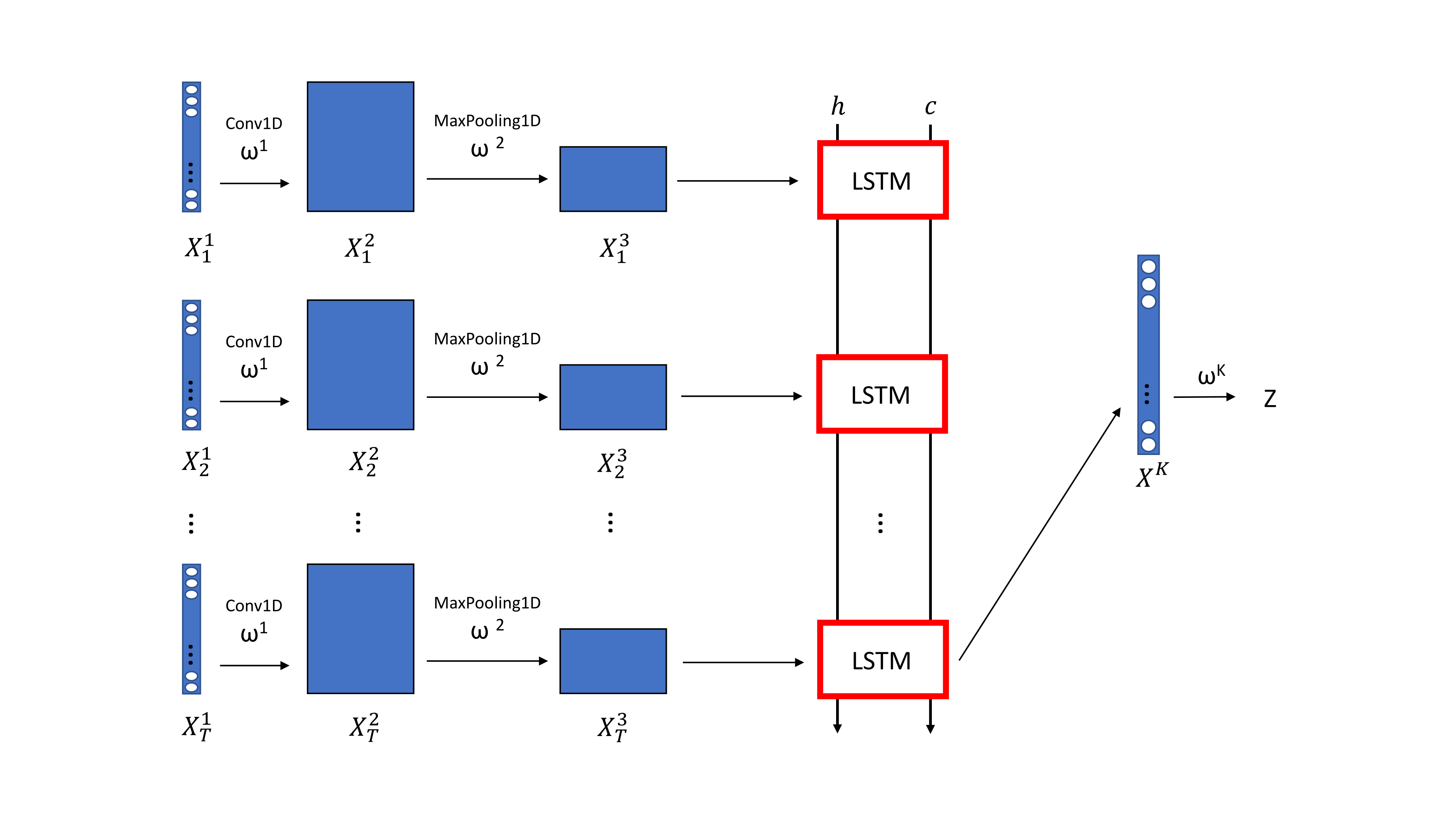}
	\caption{\label{fig:cnn_lstm_structure}CNN-LSTM Structure}
\end{figure}

In this paper, we also propose a CNN-LSTM hybrid framework to combine deep feature extraction and sequence modeling together. With deep features extracted from CNN and then processed by LSTM layers, we are able to achieve much better results, which will be discussed in details in the case study section.  As illustrated in Figure \ref{fig:cnn_lstm_structure}, the proposed CNN-LSTM model consists of two components: the first part is to implement CNN at each timestep for feature extraction and the second part is to construct a LSTM layer.


\subsection{Performance Evaluation}
In this case study, we used two performance metrics to measure how much forecasts deviate from observations. More specifically, we used accuracy to measure classification models and $R^2$ score to measure regression models. Accuracy is defined as 

\begin{equation}
	\text{Accuracy} = \frac{\text{Number of correct predictions}}{\text{Total number of predictions}}
\end{equation}

\noindent The best possible value for accuracy is 1. $R^2$ score is defined as

\begin{equation}\label{eq:r2}
	R^2 - \text{score} = 1-\frac{\sum_{i=1}^{n}(y_i-\hat{y}_i)^2}{\sum_{i=1}^{n}(y_i-\bar{y}^i)^2}
\end{equation}

\noindent where $\hat{y}_i$ is the predicted value of the $i$-th data point, $y_i$ is the actual value of the $i$-th data point, and $\bar{y}=\frac{1}{n}\sum_{i=1}^{n}y_i$. The best possible value for $R^2$ score is 1. A model that predicts $y_i$ to be a constant value $\bar{y}$ will result in a $R^2$ score of 0. The $R^2$ score can also be negative when the model performs worse than predicting everything to be $\bar{y}$.

\section{Case Study}

\subsection{Data}

The data used in this case study is taken from the Texas Department of Transportation (TxDOT)'s Pavement Management Information System (PMIS). The PMIS stores 21 flexible pavement condition indicators (e.g., rutting, cracking, patching, raveling, flushing, and roughness) as shown in Table \ref{tab:names}. Among these 21 indicators, 3 of them represent the  general condition of a road pavement: distress Score, Ride Score, and Condition Score. Distress Score reflects the amount of visible surface deterioration of a pavement. It ranges from 1 (the most distress) to 100 (the least distress). Ride Score is a measure of the pavement’s roughness, which ranges from 0.1 (the roughest) to 5.0 (the smoothest). Condition Score represents the pavement’s overall condition in terms of both distress and ride quality. It ranges from 1 (the worst condition) to 100 (the best condition). 

The rutting indicators (1-4) represent the percentage of wheelpath length with shallow, severe, failure, and deep rutting in the rated lane of the data collection section. A rut is defined as a surface depression in a wheelpath and usually indicates structural failure of the pavement. The rut depth indicators (5-7) represents the average depth of rutting measured in the wheelpath. The measurements are made by automated equipment (e.g., rut bar). The patching indicator (8) represents the percentage of lane area with patching. Patches are repairs made to pavement distress and the presence of patches indicates prior maintenance activity. The failure indicator (9) indicates the number of visually observed failures in the data collection lane. A failure is a localized part which has been severely eroded, badly cracked, or depressed. Failures can be used to identify structural deficiencies. The block cracking indicators (10) represent the percentage of lane area with block cracking in the data collection section. Block cracking consists of interconnecting cracks that divide the pavement surface into rectangular shapes with size from 1 foot by 1 foot up to 10 feet by 10 feet. Block cracking is usually caused by shrinkage of the asphalt concrete or the stablilized based courses. The alligator cracking indicator (11) represents percentage of lanes showing interconnected blocks with size less than 1 foot By 1 foot. The longitude cracking indicator (12) represents the length in feet on the data collection pavement segment. Longitudinal cracking consists of cracks  parallel to the pavement centerline. Differential movement beneath the surface is considered the primary cause of longitudinal cracking. The transverse cracking indicator (13) represents the number of visually observed transverse cracks in the rated lane. Transverse cracks are measured as the number of equivalent full lane width cracks. The raveling indicator (14) represents the degree of disintegration of the surface due to dislodgement of aggregate particles. The flushing indicator (15) indicates the degree of presence of free bitumen on the pavement surface. The International Roughness Index (IRI) indicators (16-18)  represents the rated section's longitudinal profile or ride quality. 

Among the 21 condition indicators, the raveling and flushing indicators (14, 15) are discrete variables and the rest are continuous variables. In the modeling part of this research, the discrete variables will be treated as classification problems and continuous variables will be treated as regression problems.

\singlespacing
\begin{table}[H]
\centering
\caption{\label{tab:names}Flexible Pavement Surface Condition Indicators \cite{txdot}}	
	\begin{tabular}{|p{1cm}|p{8cm}|p{3cm}|p{2cm}|p{2cm}|}
		\hline
		\textbf{\#} & \textbf{Condition Indicator} & \textbf{Explanation}                    & \textbf{Unit} & \textbf{Range}   \\ \hline
		{\color[HTML]{212529} 1}           & {\color[HTML]{212529} TX\_ACP\_RUT\_VISUAL\_SHALLOW\_PCT} & Shallow Rutting & {\color[HTML]{212529} percentage}    & {\color[HTML]{212529} 0-100}            \\ \hline
		{\color[HTML]{212529} 2}           & {\color[HTML]{212529} TX\_ACP\_RUT\_VISUAL\_SEVERE\_PCT}  & Severe Rutting & {\color[HTML]{212529} percentage}    & {\color[HTML]{212529} 0-100}            \\ \hline
		{\color[HTML]{212529} 3}           & {\color[HTML]{212529} TX\_ACP\_RUT\_VISUAL\_FAILURE\_PCT} & Failure Rutting & {\color[HTML]{212529} percentage}    & {\color[HTML]{212529} 0-100}            \\ \hline
		{\color[HTML]{212529} 4}           & {\color[HTML]{212529} TX\_ACP\_RUT\_VISUAL\_DEEP\_PCT}    & Deep Rutting & {\color[HTML]{212529} percentage}    & {\color[HTML]{212529} 0-100}            \\ \hline
		{\color[HTML]{212529} 5}           & {\color[HTML]{212529} TX\_ACP\_RUT\_LFT\_WP\_DPTH\_MEAS}  & Depth of rutting in the left wheel path & {\color[HTML]{212529} mil}           & {\color[HTML]{212529} \textgreater{}=0} \\ \hline
		{\color[HTML]{212529} 6}           & {\color[HTML]{212529} TX\_ACP\_RUT\_RIT\_WP\_DPTH\_MEAS}  & Depth of rutting in the right wheel path & {\color[HTML]{212529} mil}           & {\color[HTML]{212529} \textgreater{}=0} \\ \hline
		{\color[HTML]{212529} 7}           & {\color[HTML]{212529} TX\_ACP\_RUT\_AVG\_WP\_DEPTH\_MEAS} & Average depth of rutting in the wheel paths & {\color[HTML]{212529} mil}           & {\color[HTML]{212529} \textgreater{}=0} \\ \hline
		{\color[HTML]{212529} 8}           & {\color[HTML]{212529} TX\_ACP\_PATCHING\_PCT}             & Patching & {\color[HTML]{212529} percentage}    & {\color[HTML]{212529} 0-100}            \\ \hline
		{\color[HTML]{212529} 9}           & {\color[HTML]{212529} TX\_ACP\_FAILURE\_QTY}              & Failures & {\color[HTML]{212529} quantity}      & {\color[HTML]{212529} \textgreater{}=0} \\ \hline
		{\color[HTML]{212529} 10}          & {\color[HTML]{212529} TX\_ACP\_BLOCK\_CRACKING\_PCT}      & Block cracking & {\color[HTML]{212529} percentage}    & {\color[HTML]{212529} 0-100}            \\ \hline
		{\color[HTML]{212529} 11}          & {\color[HTML]{212529} TX\_ACP\_ALLIGATOR\_CRACKING\_PCT}  & Alligator cracking & {\color[HTML]{212529} percentage}    & {\color[HTML]{212529} 0-100}            \\ \hline
		{\color[HTML]{212529} 12}          & {\color[HTML]{212529} TX\_ACP\_LONGITUDE\_CRACKING\_PCT}  & Longitude cracking & {\color[HTML]{212529} foot}          & {\color[HTML]{212529} \textgreater{}=0} \\ \hline
		{\color[HTML]{212529} 13}          & {\color[HTML]{212529} TX\_ACP\_TRANSVERSE\_CRACKING\_QTY} & Transverse cracking & {\color[HTML]{212529} quantity}      & {\color[HTML]{212529} \textgreater{}=0} \\ \hline
		{\color[HTML]{212529} 14}          & {\color[HTML]{212529} TX\_ACP\_RAVELING\_CODE}            & Raveling & {\color[HTML]{212529} level}         & {\color[HTML]{212529} 1,2,3,4}          \\ \hline
		{\color[HTML]{212529} 15}          & {\color[HTML]{212529} TX\_ACP\_FLUSHING\_CODE}            & Flushing & {\color[HTML]{212529} level}         & {\color[HTML]{212529} 1,2,3,4}          \\ \hline
		{\color[HTML]{212529} 16}          & {\color[HTML]{212529} TX\_IRI\_LEFT\_SCORE}               & Left IRI & {\color[HTML]{212529} in/mile}       & {\color[HTML]{212529} \textgreater{}=0} \\ \hline
		{\color[HTML]{212529} 17}          & {\color[HTML]{212529} TX\_IRI\_RIGHT\_SCORE}              & Right IRI & {\color[HTML]{212529} in/mile}       & {\color[HTML]{212529} \textgreater{}=0} \\ \hline
		{\color[HTML]{212529} 18}          & {\color[HTML]{212529} TX\_IRI\_AVERAGE\_SCORE}            & Avg. IRI & {\color[HTML]{212529} in/mile}       & {\color[HTML]{212529} \textgreater{}=0} \\ \hline
		{\color[HTML]{212529} 19}          & {\color[HTML]{212529} TX\_RIDE\_SCORE}                    & Ride Score & {\color[HTML]{212529} -}             & {\color[HTML]{212529} 0-100}            \\ \hline
		{\color[HTML]{212529} 20}          & {\color[HTML]{212529} TX\_DISTRESS\_SCORE}                & Distress Score & {\color[HTML]{212529} -}             & {\color[HTML]{212529} 0-100}            \\ \hline
		{\color[HTML]{212529} 21}          & {\color[HTML]{212529} TX\_CONDITION\_SCORE}               & Condition Score & {\color[HTML]{212529} -}             & {\color[HTML]{212529} 0-100}            \\ \hline
	\end{tabular}
\end{table}
\singlespacing

\begin{figure}[H]
	\centering
	\includegraphics[width=1\textwidth]{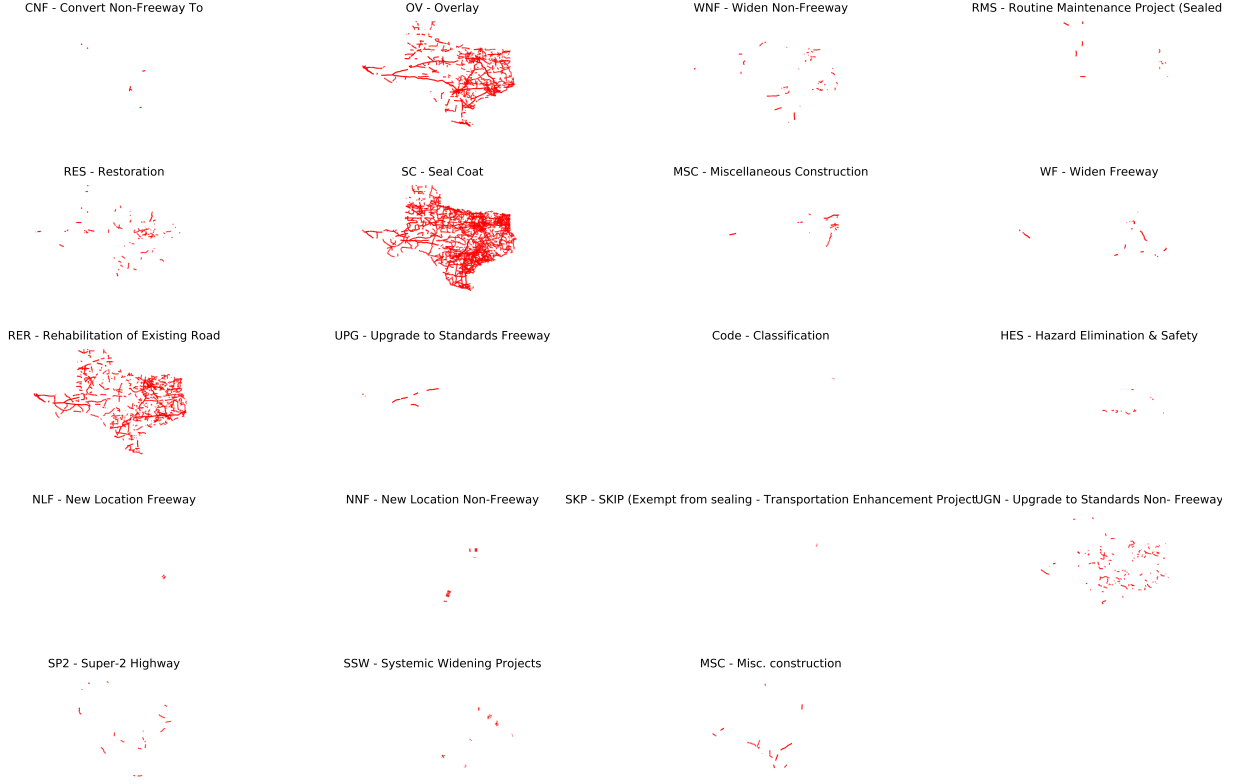}
	\caption{\label{fig:mn_map}Maintenance \& Rehabilitation Work Map}
\end{figure}

In this case study, we also collected the road work history data. For each pavement section (around 0.5 mile length), we collected data indicating the type of last road work and the time it was implemented \citep{gao2021detection}. Table \ref{tab:mn_number} shows 20 different projects and the corresponding number of data points. The top road work in the dataset include seal coat, overlay, and rehabilitation of existing road, which are common preventive maintenance and rehabilitation methods used by TxDOT \citet{gharaibeh2012evaluation}. In Table \ref{tab:mn_number}, rehabilitation of existing road include reshaping, resurfacing, and addition of existing base. Super-2 highway refers to construction of periodic passing lane to a 2 lane rural highway to allow passing of slower vehicles and the dispersal of traffic platoons. Miscellaneous construction includes signing, pavement markings, illumination, adding turn lanes, and adding or moving entrance or exit ramps. The popularity of these three treatments can also be found in Figure \ref{fig:mn_map}, which shows the location where the treatments were implemented.

\begin{table}[H]
	\caption{\label{tab:mn_number}Number of Data Points by Road Work Type}
	\begin{tabular}{|c|c|l|c|}
		\hline
	\textbf{\#} &	{\textbf{Work Code}}          & {\color[HTML]{212529} \textbf{Work Description}}                                            & {\color[HTML]{212529} \textbf{Count}}          \\ \hline
	1           &	                   & {\color[HTML]{212529} Do Nothing}                                                           & {\color[HTML]{212529} 47217}   \\ \hline
	2           &	{\color[HTML]{212529} 9}                  & {\color[HTML]{212529} SC - Seal Coat}                                                       & {\color[HTML]{212529} 32763}          \\ \hline
	3           &	{\color[HTML]{212529} 12}                 & {\color[HTML]{212529} RER - Rehabilitation of Existing Road}                                & {\color[HTML]{212529} 9308}           \\ \hline
 	4           &	{\color[HTML]{212529} 4}                  & {\color[HTML]{212529} OV - Overlay}                                                         & {\color[HTML]{212529} 5759}           \\ \hline
	5           &	{\color[HTML]{212529} 9}                  & {\color[HTML]{212529} P05 - Full Width Seal Coat}                                           & {\color[HTML]{212529} 3779}           \\ \hline
	6           &	{\color[HTML]{212529} 44}                 & {\color[HTML]{212529} MSC - Miscellaneous construction}                                     & {\color[HTML]{212529} 2163}           \\ \hline
	7           &	{\color[HTML]{212529} 7}                  & {\color[HTML]{212529} RES - Restoration}                                                    & {\color[HTML]{212529} 985}            \\ \hline
	8           &	{\color[HTML]{212529} 11}                 & {\color[HTML]{212529} WF - Widen Freeway}                                                   & {\color[HTML]{212529} 965}            \\ \hline
	9           &	{\color[HTML]{212529} 40}                 & {\color[HTML]{212529} SP2 - Super-2 Highway}                                                & {\color[HTML]{212529} 765}            \\ \hline
	10           &	{\color[HTML]{212529} 13}                 & {\color[HTML]{212529} UPG - Upgrade to Standards Freeway}                                   & {\color[HTML]{212529} 575}            \\ \hline
	11           &	{\color[HTML]{212529} 5}                  & {\color[HTML]{212529} WNF - Widen Non-Freeway}                                              & {\color[HTML]{212529} 462}            \\ \hline
	12           &	{\color[HTML]{212529} 38}                 & {\color[HTML]{212529} UGN - Upgrade to Standards Non- Freeway}                              & {\color[HTML]{212529} 447}            \\ \hline
	13 &	{\color[HTML]{212529} 10}                 & {\color[HTML]{212529} MSC - Miscellaneous Construction}                                     & {\color[HTML]{212529} 228}            \\ \hline
	14 &	{\color[HTML]{212529} 22}                 & {\color[HTML]{212529} HES - Hazard Elimination \& Safety}                                   & {\color[HTML]{212529} 111}            \\ \hline
	15 &	{\color[HTML]{212529} 41}                 & {\color[HTML]{212529} SSW - Systemic Widening Projects}                                     & {\color[HTML]{212529} 98}             \\ \hline
	16 &	{\color[HTML]{212529} 28}                 & {\color[HTML]{212529} NNF - New Location Non-Freeway}                                       & {\color[HTML]{212529} 94}             \\ \hline
	17 &	{\color[HTML]{212529} 6}                  & {\color[HTML]{212529} RMS - Routine Maintenance Project (Sealed)}                           & {\color[HTML]{212529} 50}             \\ \hline
	18 &	{\color[HTML]{212529} 2}                  & {\color[HTML]{212529} CNF - Convert Non-Freeway To}                                         & {\color[HTML]{212529} 48}             \\ \hline
	19 &	{\color[HTML]{212529} 33}                 & {\color[HTML]{212529} SKP - SKIP - Transportation Enhancement Project} & {\color[HTML]{212529} 23}             \\ \hline
	20 &	{\color[HTML]{212529} 27}                 & {\color[HTML]{212529} NLF - New Location Freeway}                                           & {\color[HTML]{212529} 8}              \\ \hline
	\end{tabular}
\end{table}

In this case study, the historical pavement management data collected between 2000 and 2018 are used to build pavement performance models. Each data point consists of 378 feature variables (21 indicators multiplied by 17 years, 20 M\&R dummy variables, and 1 continuous variable indicating the time since last treatment). We developed a performance model for each of the 21 condition indicators with the target variable being the condition value in 2018. 20\% of the dataset is used for testing and the rest for training the models. 



\subsection{Model Structure}

Deep learning models consist of parameterized functions and the selection of parameters have direct impact on the performance results. To find out the optimal values for parameters such as number of filters, size of filters, and number of layers, we have tested on various configurations for CNN, LSTM and CNN-LSTM models. The model structures are presented in Tables \ref{tab:cnn}, \ref{tab:lstm}, and \ref{tab:cnn_lstm}. For example, for the CNN model, the hidden layer contains 3 convolution layers with filters 32, 64, and 128. Each convolutional layer is followed by maxpooling2D with the pool length 2. 

Unlike traditional machine learning models, deep learning models don't need any engineering and domain expertise to design a feature extractor. Instead, the features are learned from the raw data using a general-purpose learning process. In this paper, our application is on analyzing pavement condition indicators over a certain time period and the data is constructed to be two dimensional. In the input layer of the models, each data point is reshaped to a 2D "image" and the input matrix has the structure shown in Equation \eqref{eq:cnn_matrix}.  

\begin{equation}\label{eq:cnn_matrix}
\begin{pmatrix}
x_{2000,1} & x_{2000,2} & \dots & x_{2000,21} & s_{2000} & m_1 & \dots & m_{20} \\
x_{2001,1} & x_{2001,2} & \dots & x_{2001,21} & s_{2001} & m_1 & \dots & m_{20} \\
\vdots & \vdots & \ddots & \vdots & \vdots & \vdots & \vdots & \vdots \\
x_{2017,1} & x_{2017,2} & \dots & x_{2017,21} & s_{2018} & m_1 & \dots & m_{20} \\
\end{pmatrix}
\end{equation}

\noindent where $x_{t,i}$ represents the $i$th indicator value collected in the $t$th year. $s_t$ represents the number of years between the time of the last treatment and year $t$. $m_j$ indicates the existence of the implementation of the $j$ the treatment.

\begin{table}[H]
		\centering
	\caption{\label{tab:cnn}Structure and Configuration Details of the Proposed CNN Model}

	\begin{threeparttable}	
	\begin{tabular}{ |c|c|c|c|c|c|c|} 
		\hline
		Layer & Type & Output Shape & Parameters & Filters & Kernel-size & Pool-size \\ 
		\hline
		& Input & (18, 42, 1) & - & - & - & - \\   
		1 & Conv2D & (18, 42, 32) & 320 & 32 & 3 & - \\  
		2 & MaxPooling2D & (9, 21, 32)  & 0 & - & - & 2 \\ 
		3 & Dropout(0.25) & (9, 21, 32) & 0 & - & - & - \\ 
		4 & Conv2D & (9, 21, 64)  & 18,496  & 64  & 3 & - \\   
		5 & MaxPooling2D & (5, 11, 64) & 0 & - & - & 2 \\  
		6 & Dropout(0.25) & (5, 11, 64) & 0 & - & - & - \\
		7 & Conv2D & (5,11,128) & 73,856   & 128  & 3 & - \\   
		8 & MaxPooling2D & (3, 6, 128) & 0 & - & - & 2 \\  
		9 & Dropout(0.4) & (3, 6, 128) & 0 & - & - & - \\   
		10 & Fully-connected (128) & (128)  & 295,040   & - & - & - \\    
		11 & Dropout(0.3) & (128) & 0 & - & - & - \\    
		12 & Fully-connected & (1)$^1$ or (4)$^2$  & 129$^1$ or 516$^2$    & - & - & - \\     
		\hline
	\end{tabular}
	\begin{tablenotes}
		\item[1] For TX\_ACP\_RAVELING\_CODE and TX\_ACP\_FLUSHING\_CODE.
		\item[2] For other condition indicators.
	\end{tablenotes}
	\end{threeparttable}
\end{table}

The LSTM model configuration details are provided in Table \eqref{tab:lstm}. The input to the LSTM model is the same as in Equation \eqref{eq:cnn_matrix}. The hidden layer contains 50 memory blocks. The fully connected layer is followed by output layer with sigmoid non-linear activation function. 

\begin{table}[H]
	\centering
	\caption{\label{tab:lstm}Structure and Configuration Details of the Proposed LSTM Model}
		
\begin{threeparttable}	

	\begin{tabular}{ |c|c|c|c| } 
		\hline
		Layer & Type & Output Shape & Parameters \\ 
		\hline
		& Input & (18, 42) & -  \\   
		1 & LSTM (50) & (50) & 18600  \\  
		2 & Fully-connected & (1)$^1$ or (4)$^2$ & 51$^1$ or 204$^2$ \\    
		\hline
	\end{tabular}
\begin{tablenotes}
	\item[1] For TX\_ACP\_RAVELING\_CODE and TX\_ACP\_FLUSHING\_CODE.
	\item[2] For other condition indicators.
\end{tablenotes}
\end{threeparttable}	
\end{table}

The input layer of the CNN-LSTM model is the same as in Equation \eqref{eq:cnn_matrix}. Each of the rows of the input matrix will go through a convolution layer with 32 filters followed by maxpooling1D with the pool length of 2. Finally, the features learnt by CNN network will be passed to the LSTM layer, which contains 50 memory blocks. 

\begin{table}[H]
	\caption{\label{tab:cnn_lstm}Structure and Configuration Details of the Proposed CNN+LSTM Model}
	\centering
\begin{threeparttable}		
	\begin{tabular}{ |p{1cm}|p{4cm}|p{2cm}|p{2cm}|p{2cm}|p{2cm}|p{2cm}| } 
		\hline
		Layer & Type & Output Shape & Parameters & Filters & Kernel-size & Pool-size \\ 
		\hline
		& Input & (18, 42, 1) & - & - & - & -  \\    
		\hline
		1 & TimeDistributed (Conv1D)  & (18, 42, 32) & 128 & 32 & 3 & - \\  
		\hline
		2 & Timedistributed (MaxPooling1D) & (18, 21, 32) & 0 & - & - & 2 \\ 
		\hline
		3 & Timedistributed (Flatten) & (18, 672) & 0 & - & - & - \\ 
		\hline 
		3 & LSTM (50) & (50) & 144,600  & - & - & - \\   
		\hline
		5 & Fully-connected & (1)$^1$ or (4)$^2$ & 51$^1$ or 204$^2$  & - & - & -  \\    
		\hline
	\end{tabular}
\begin{tablenotes}
	\item[1] For TX\_ACP\_RAVELING\_CODE and TX\_ACP\_FLUSHING\_CODE.
	\item[2] For other condition indicators.
\end{tablenotes}
\end{threeparttable}
\end{table}

\subsection{Results}

The modeling results are provided in Table \ref{tab:deep_results}. For each of the condition indicator, there are more than 100,000 data points (pavement sections). The difference in the number of data points are mainly due to the availability of data in specific years. Overall, the CNN models perform better than LSTM and CNN+LSTM models for more than half of the condition indicators. However, there are only five models (rut failure, ride score, left IRI, right IRI, and average IRI) have $R^2$ scores greater than 0.7. This results show that roughness related indicators have better modeling results than other distress indicators. This is probably because the roughness data collection process is more consistent and reliable than that of other distress data. The results indicate that the developed roughness related models are suitable for network-level predictions. 

It is also noteworthy to mention that there are several indicators (e.g., severe rut, patching, failure, block cracking, and transverse cracking) with large negative $R^2$ score values. This indicates that the data cannot explain the target variable and the models perform poorly at predicting the testing set. It is primarily because the target variables have little variation in the dataset, which results in very small values of sum-of-squares from the horizontal line (the denominator in Equation \ref{eq:r2}). 

\singlespacing
\begin{table}[H]
		\caption{\label{tab:deep_results}Summary of Model Performance (on Testing Set) at the 100th Epoch}
	\footnotesize
	\begin{threeparttable}		
	\begin{tabular}{|l|p{2cm}|p{2cm}|p{2cm}|p{2cm}|}
		\hline
		{\color[HTML]{212529} Condition Indicator}                          & {\color[HTML]{212529} CNN (r2 score or accuracy)} & {\color[HTML]{212529} LSTM (r2 score or accuracy)} & {\color[HTML]{212529} CNN+LSTM (r2 score or accuracy)} & {\color[HTML]{212529} Number of Data Points} \\ \hline
		{\color[HTML]{212529} TX\_ACP\_RUT\_VISUAL\_SHALLOW\_PCT} & {\color[HTML]{212529} 0.55}                       & {\color[HTML]{212529} 0.51}                        & {\color[HTML]{212529} 0.57\textsuperscript{*}}                            & {\color[HTML]{212529} 101,730}               \\ \hline
		{\color[HTML]{212529} TX\_ACP\_RUT\_VISUAL\_DEEP\_PCT}    & {\color[HTML]{212529} -0.02}                      & {\color[HTML]{212529} 0.07\textsuperscript{*}}                        & {\color[HTML]{212529} -20547.22}                       & {\color[HTML]{212529} 101,730}               \\ \hline
		{\color[HTML]{212529} TX\_ACP\_RUT\_VISUAL\_SEVERE\_PCT}  & {\color[HTML]{212529} -336800.48}                 & {\color[HTML]{212529} -214955.76\textsuperscript{*}}                  & {\color[HTML]{212529} -528423.16}                      & {\color[HTML]{212529} 101,730}               \\ \hline
		{\color[HTML]{212529} TX\_ACP\_RUT\_VISUAL\_FAILURE\_PCT} & {\color[HTML]{212529} 0.7\textsuperscript{*}}                        & {\color[HTML]{212529} -217.95}                     & {\color[HTML]{212529} -97.92}                          & {\color[HTML]{212529} 101,730}               \\ \hline
		{\color[HTML]{212529} TX\_ACP\_RUT\_LFT\_WP\_DPTH\_MEAS}  & {\color[HTML]{212529} 0.45\textsuperscript{*}}                       & {\color[HTML]{212529} 0.32}                        & {\color[HTML]{212529} 0.33}                            & {\color[HTML]{212529} 101,699}               \\ \hline
		{\color[HTML]{212529} TX\_ACP\_RUT\_RIT\_WP\_DPTH\_MEAS}  & {\color[HTML]{212529} 0.47\textsuperscript{*}}                       & {\color[HTML]{212529} 0.37}                        & {\color[HTML]{212529} 0.34}                            & {\color[HTML]{212529} 101,699}               \\ \hline
		{\color[HTML]{212529} TX\_ACP\_RUT\_AVG\_WP\_DEPTH\_MEAS} & {\color[HTML]{212529} 0.52\textsuperscript{*}}                       & {\color[HTML]{212529} 0.44}                        & {\color[HTML]{212529} 0.48}                            & {\color[HTML]{212529} 101,699}               \\ \hline
		{\color[HTML]{212529} TX\_ACP\_PATCHING\_PCT}             & {\color[HTML]{212529} -9851665.16}                & {\color[HTML]{212529} -1364742.29}                 & {\color[HTML]{212529} -1.01\textsuperscript{*}}                           & {\color[HTML]{212529} 101,754}               \\ \hline
		{\color[HTML]{212529} TX\_ACP\_FAILURE\_QTY}              & {\color[HTML]{212529} -652165.62\textsuperscript{*}}                 & {\color[HTML]{212529} -657126.63}                  & {\color[HTML]{212529} -32008948.61}                    & {\color[HTML]{212529} 101,754}               \\ \hline
		{\color[HTML]{212529} TX\_ACP\_BLOCK\_CRACKING\_PCT}      & {\color[HTML]{212529} -43009529.2\textsuperscript{*}}                & {\color[HTML]{212529} -47330983.02}                & {\color[HTML]{212529} -19679668.25}                    & {\color[HTML]{212529} 101,754}               \\ \hline
		{\color[HTML]{212529} TX\_ACP\_ALLIGATOR\_CRACKING\_PCT}  & {\color[HTML]{212529} -1.42}                      & {\color[HTML]{212529} -0.67\textsuperscript{*}}                       & {\color[HTML]{212529} -3.34}                           & {\color[HTML]{212529} 101,754}               \\ \hline
		{\color[HTML]{212529} TX\_ACP\_LONGITUDE\_CRACKING\_PCT}  & {\color[HTML]{212529} 0.24\textsuperscript{*}}                       & {\color[HTML]{212529} -0.24}                       & {\color[HTML]{212529} -0.37}                           & {\color[HTML]{212529} 101,754}               \\ \hline
		{\color[HTML]{212529} TX\_ACP\_TRANSVERSE\_CRACKING\_QTY} & {\color[HTML]{212529} -486099.71}                 & {\color[HTML]{212529} -411265.58\textsuperscript{*}}                  & {\color[HTML]{212529} -1696195.16}                     & {\color[HTML]{212529} 101,754}               \\ \hline
		{\color[HTML]{212529} TX\_ACP\_RAVELING\_CODE \textsuperscript{\#}}            & {\color[HTML]{212529} 0.57}                       & {\color[HTML]{212529} 0.56}                        & {\color[HTML]{212529} 0.59\textsuperscript{*}}                            & {\color[HTML]{212529} 101,754}               \\ \hline
		{\color[HTML]{212529} TX\_ACP\_FLUSHING\_CODE \textsuperscript{\#}}            & {\color[HTML]{212529} 0.47}                       & {\color[HTML]{212529} 0.48\textsuperscript{*}}                        & {\color[HTML]{212529} 0.46}                            & {\color[HTML]{212529} 101,754}               \\ \hline
		{\color[HTML]{212529} TX\_CONDITION\_SCORE}               & {\color[HTML]{212529} 0.42\textsuperscript{*}}                       & {\color[HTML]{212529} 0.11}                        & {\color[HTML]{212529} -0.1}                            & {\color[HTML]{212529} 105,634}               \\ \hline
		{\color[HTML]{212529} TX\_DISTRESS\_SCORE}                & {\color[HTML]{212529} 0.3\textsuperscript{*}}                        & {\color[HTML]{212529} -0.06}                       & {\color[HTML]{212529} -0.08}                           & {\color[HTML]{212529} 105,717}               \\ \hline
		{\color[HTML]{212529} TX\_RIDE\_SCORE}                    & {\color[HTML]{212529} 0.68}                       & {\color[HTML]{212529} 0.78}                        & {\color[HTML]{212529} 0.79\textsuperscript{*}}                            & {\color[HTML]{212529} 105,696}               \\ \hline
		{\color[HTML]{212529} TX\_IRI\_LEFT\_SCORE}               & {\color[HTML]{212529} 0.7}                        & {\color[HTML]{212529} 0.73\textsuperscript{*}}                        & {\color[HTML]{212529} -0.08}                           & {\color[HTML]{212529} 105,696}               \\ \hline
		{\color[HTML]{212529} TX\_IRI\_RIGHT\_SCORE}              & {\color[HTML]{212529} 0.75\textsuperscript{*}}                       & {\color[HTML]{212529} 0.72}                        & {\color[HTML]{212529} -0.12}                           & {\color[HTML]{212529} 105,696}               \\ \hline
		{\color[HTML]{212529} TX\_IRI\_AVERAGE\_SCORE}            & {\color[HTML]{212529} 0.76\textsuperscript{*}}                       & {\color[HTML]{212529} 0.7}                         & {\color[HTML]{212529} -0.12}                           & {\color[HTML]{212529} 105,696}               \\ \hline
	\end{tabular}
	\begin{tablenotes}
		\item[\#] Accuracy is used as model performance metric.
		\item[*] Best model.
	\end{tablenotes}
	\end{threeparttable}
\end{table}

\singlespacing

\begin{figure}[H]
	\centering
	\includegraphics[width=1\textwidth]{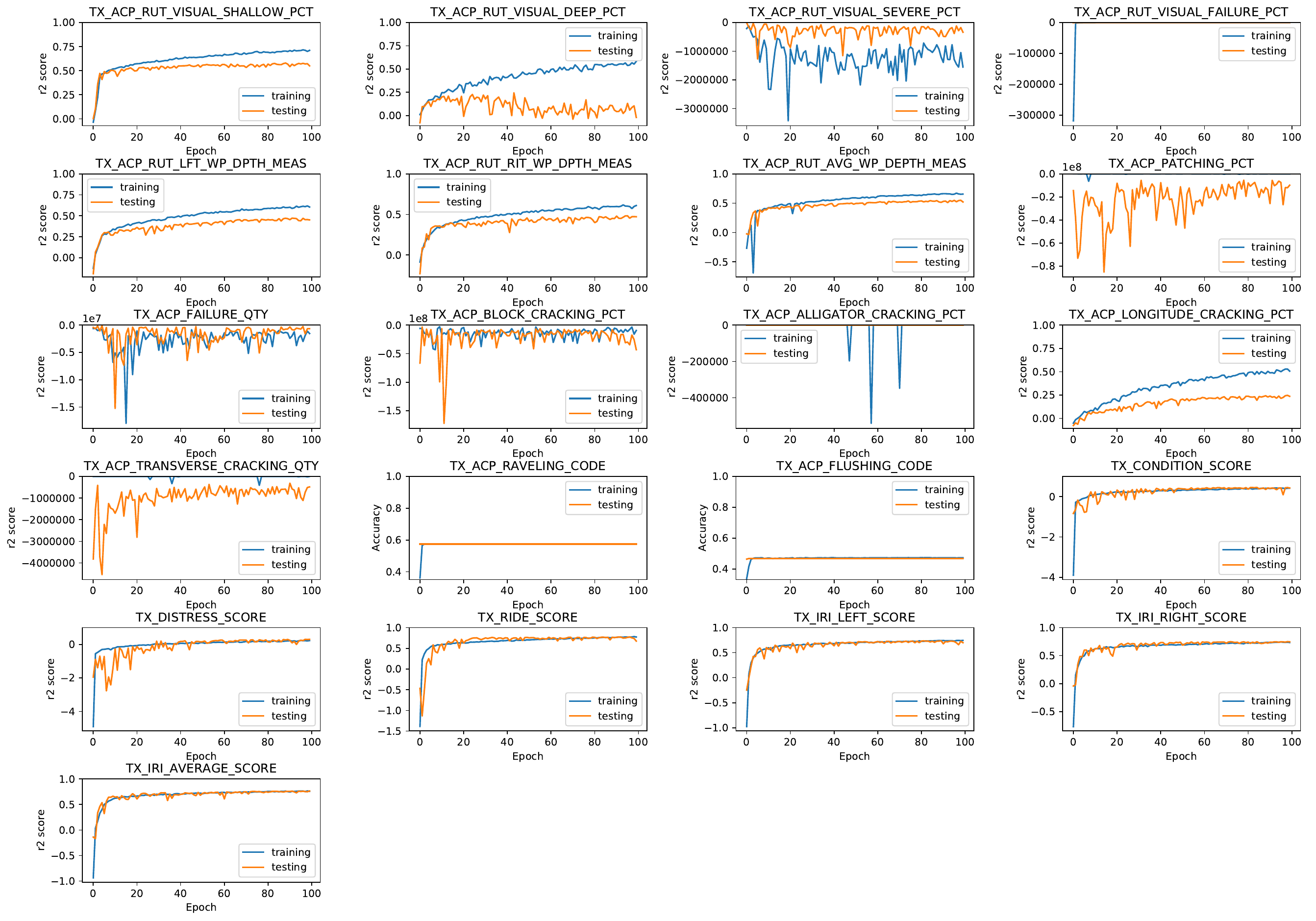}
	\caption{\label{fig:epoch}CNN training history: $R^2$ Score or Accuracy v.s. Epoch}
\end{figure}

Figure \eqref{fig:epoch} shows the training history of the CNN models. While the plots of raveling and flushing represent the accuracy of the models, the rest of the plots show the $R^2$ score over the training epochs. The history for the training dataset is labled as training and the history for the testing dataset is labeled as testing. From the plots we can see that the models for condition score, distress score, ride score, left IRI, right IRI, and average IRI could probably achieve higher value of $R^2$ score if trained a little more epochs as the trend for $R^2$ score on these datasets is still rising for the last few epochs. It can be found that the models for the indicators mentioned above has not yet over-learned the training dataset, showing comparable performance on both datasets.

\begin{figure}[H]
	\centering
	\includegraphics[width=1\textwidth]{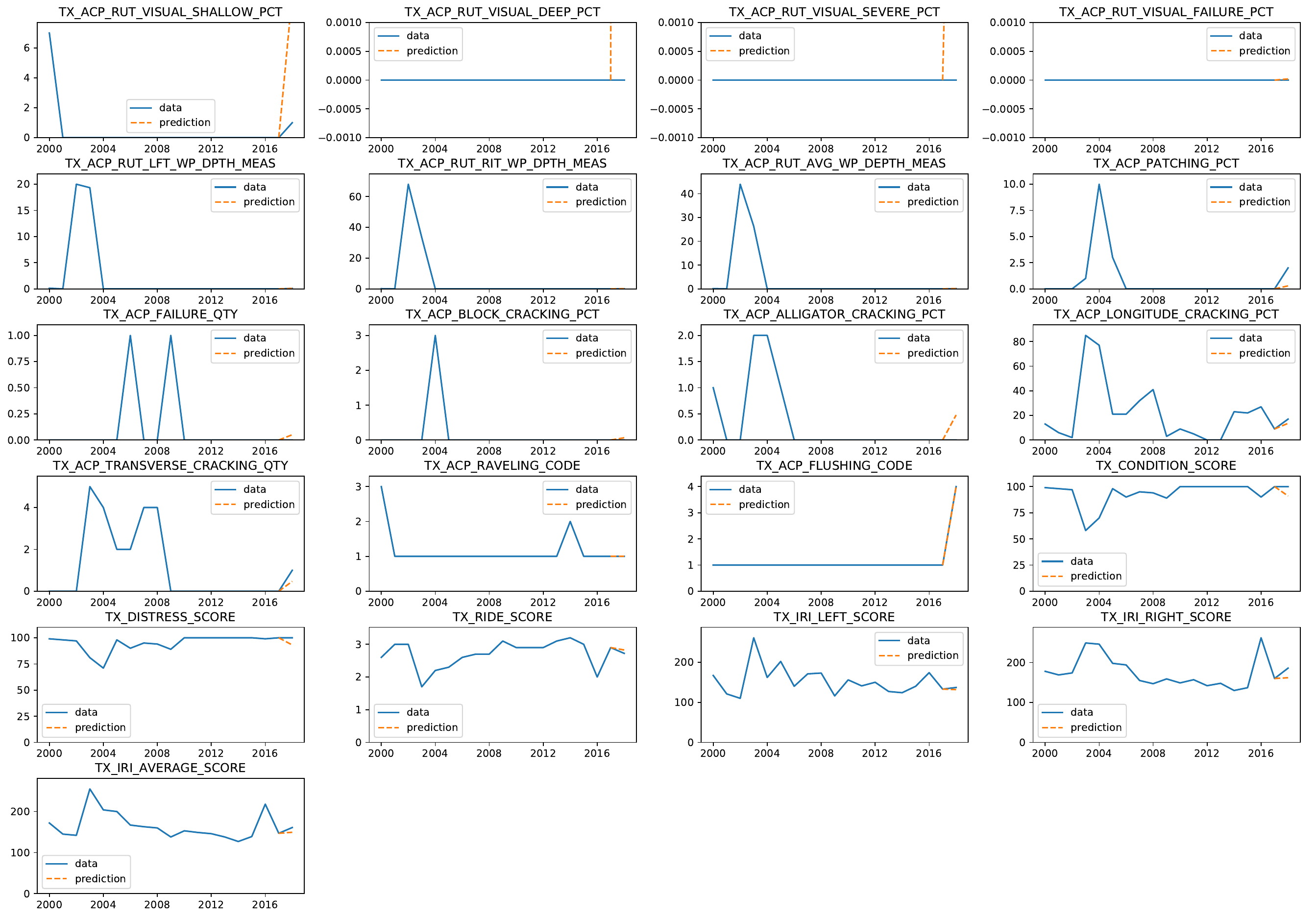}
	\caption{\label{fig:predicted}Actual Data vs. Predicted Value}
\end{figure}

Figure \eqref{fig:predicted} shows the plots of actual condition value and predicted value in 2018 using the the CNN model results of a randomly selected pavement section. As can be seen in the figure, the models of rut failure, rut depth, logitude cracking, raveling, flusing, ride score, left IRI, and average IRI are able to produce reasonable predictions. For indicators where historical data remain constant (e.g., rut deep, rut severe), the predicted values are quite different from the true values. Combining this with the results from Table \ref{tab:deep_results}, the developed models are able to produce reasonable results for roughness related indicators at project level.

\section{Conclusions}
In this paper, we used deep learning models including CNN, LSTM, and CNN-LSTM to analyze pavement condition data for performance prediction. The results achieved $R^2$ scores greater than 0.70 for a number of condition indicators. A total of 378 condition attributes including rutting, cracking, raveling, flushing, and roughness were used in the modeling process. Moreover, work history of 20 types of maintenance treatments were also taken into consideration in the proposed model. The proposed deep learning models were calibrated with the TxDOT PMIS data. The case study results confirmed the effectiveness and robustness of the proposed model. The developed model can assist engineers and administrators in more effectively managing infrastructure systems through improved performance prediction; in particular, it can help budget allocation and needs analysis at the network level. Furthermore, although the case study in this paper was developed and tested for pavement deterioration, it can be easily implemented and extended to characterize the performance of other infrastructure facilities.

\section*{Data Availability Statement}
Because of a non-disclosure agreement with the sponsor (TxDOT), pavement condition and maintenance work data used in this research is not available to the public. But all of the models and code that support the findings of this study are available from the corresponding author upon reasonable request.

\section*{Acknowledgement}
This work was supported by the Texas Department of Transportation Grant 0-6988. The authors would like to thank all Texas Department of Transportation personnel, who have helped this research study. All opinions, errors, omissions, and recommendations in this paper are the responsibility of the authors.


\bibliographystyle{unsrtnat}
\bibliography{ref}

\end{document}